\documentclass[sigconf, screen]{acmart} 

\usepackage{multirow} 
\usepackage[table,dvipsnames]{xcolor}
\definecolor{navyblue}{HTML}{0071BC}

\usepackage{tcolorbox} 
\newcommand{\hypbox}[2]{%
\begin{tcolorbox}[colback=white!98!black,colframe=white!30!black,boxsep=1.1pt,top=6.75pt]%
\vspace{1.75pt}%
\textbf{#1}\\[-0.575em]
\noindent\makebox[\textwidth]{\rule{\textwidth}{0.4pt}}
\\[0.25em]
#2
\end{tcolorbox}
}

\AtBeginDocument{%
  }

\setcopyright{acmlicensed}
\copyrightyear{2018}
\acmYear{2026}
\acmDOI{XXXXXXX.XXXXXXX}
\acmConference[Conference acronym 'XX]{Make sure to enter the correct conference title from your rights confirmation email}{June 03--05, 2018}{Woodstock, NY}

\acmISBN{978-1-4503-XXXX-X/2018/06}

\begin{document}

\title{ScalSelect: Scalable Training-Free Multimodal Data Selection for Efficient Visual Instruction Tuning}


\author{
  Changti Wu\textsuperscript{\rm 1,\rm 2},\space 
  Jiahuai Mao\textsuperscript{\rm 3},\space
  Yuzhuo Miao\textsuperscript{\rm 4,\rm 2},\space 
  Shijie Lian\textsuperscript{\rm 5,\rm 2},\space
  Bin Yu\textsuperscript{\rm 4,\rm 2},\space 
  Xiaopeng Lin\textsuperscript{\rm 6,\rm 7},\\ 
  Cong Huang\textsuperscript{\rm 2,\rm 7},\space 
  Lei Zhang\textsuperscript{\rm 1}\textsuperscript{$\dag$},\space
  Kai Chen\textsuperscript{\rm 2,\rm 7}\textsuperscript{$\dag$}
}

\affiliation{%
  \institution{\textsuperscript{1}East China Normal University\space
    \textsuperscript{2}Zhongguancun Academy\space
    \textsuperscript{3}The Hong Kong Polytechnic University\\
    \textsuperscript{4}Harbin Institute of Technology\space
    \textsuperscript{5}Huazhong University of Science and Technology\\
    \textsuperscript{6}The Hong Kong University of Science and Technology (Guangzhou)\\
    \textsuperscript{7}Zhongguancun Institute of Artificial Intelligence} 
  \country{}
}
\thanks{$\dag$ Corresponding author}

\begin{abstract}
Large-scale Visual Instruction Tuning (VIT) has become a key paradigm for advancing the performance of vision-language models (VLMs) across various multimodal tasks. However, training on the large-scale datasets is computationally expensive and inefficient due to redundancy in the data, which motivates the need for multimodal data selection to improve training efficiency.
Existing data selection methods for VIT either require costly training or gradient computation.
Training-free alternatives often depend on proxy models or datasets, instruction-agnostic representations, and pairwise similarity with quadratic complexity, limiting scalability and representation fidelity.
In this work, we propose ScalSelect, a scalable training-free multimodal data selection method with linear-time complexity with respect to the number of samples, eliminating the need for external models or auxiliary datasets.
ScalSelect first constructs sample representations by extracting visual features most attended by instruction tokens in the target VLM, capturing instruction-relevant information. It then identifies samples whose representations best approximate the dominant subspace of the full dataset representations, enabling scalable importance scoring without pairwise comparisons.
Extensive experiments across multiple VLMs, datasets, and selection budgets demonstrate that ScalSelect achieves over 97.5\% of the performance of training on the full dataset using only 16\% of the data, and even outperforms full-data training in some settings.
The code is available at \href{https://github.com/ChangtiWu/ScalSelect}{ScalSelect}.
\end{abstract}  






\settopmatter{printacmref=false} 
\renewcommand\footnotetextcopyrightpermission[1]{} 
\pagestyle{plain} 
\setcopyright{none}

\maketitle

\section{Introduction}
    



Large-scale Visual Instruction Tuning (VIT) \cite{liu2023visual,liu2024improved,zhao2023svit} has become a cornerstone for advancing the capabilities of modern vision-language models (VLMs) \cite{zhang2024vision,ghosh2024exploring,li2025survey}. By leveraging massive collections of image-instruction pairs, recent VLMs have demonstrated strong performance across a wide range of multimodal tasks, including visual question answering, optical character recognition (OCR), diagram understanding, and multimodal reasoning \cite{liu2024llavanext,wu2024deepseek,wei2025deepseek,liu2026chartverse,wu2025dynasolidgeo,lian2025euclid}. However, as the scale of visual instruction datasets continues to grow, full fine-tuning on the entire dataset becomes increasingly expensive and computationally inefficient \cite{shinde2025survey}. In practice, such datasets often contain substantial redundancy, where many samples are highly overlapping in content or supervision and contribute limited marginal gains while incurring significant computational cost. This motivates the need for effective data selection methods that can identify a compact subset of samples, enabling efficient training while maintaining the competitive performance of full fine-tuning.

To address data selection, a number of existing approaches adopt training-based or gradient-based strategies. Training-based methods \cite{liu2025less,wu2024icons,ma2025mllm,chen2024your} typically rely on warm-up training or training a proxy model to estimate sample importance, while gradient-based methods \cite{yang2024clip,maharana2024adapt} require backpropagation through the target model to measure the influence of samples. Although effective in certain settings, these approaches inevitably introduce additional training procedures and computational overhead, partially conflicting with the original goal of data selection. Moreover, the reliance on proxy models or auxiliary training stages complicates deployment and limits scalability, especially for large-scale multimodal datasets. As a result, there has been growing interest in training-free data selection methods that aim to avoid the high computational cost associated with training or backpropagation while still preserving model performance.

Existing training-free multimodal data selection methods \cite{lee2024concept,bi2025prism,ivison2025large,tang2025training} often rely on the similarity-related metrics between sample representations. Despite efficiency advantages, they often suffer from limitations in how sample representations are constructed. Many approaches either rely on external proxy models or datasets, such as CLIP-based visual representations, or aggregate all visual tokens from the target model to form a sample representation. These strategies are largely instruction-agnostic: they treat visual content independently of the accompanying textual instruction. However, in visual instruction tuning, the semantic meaning of a sample is largely conditioned on the instruction. The same image may require attention to entirely different visual regions under different instructions. By indiscriminately aggregating all visual tokens without considering instruction relevance, existing methods risk mixing instruction-relevant and irrelevant information, resulting in sample representations failing to distinguish such cases effectively.

Another fundamental limitation shared by many prior methods lies in their local perspective on data selection. Most existing approaches \cite{lee2024concept,bi2025prism,ivison2025large,tang2025training,liu2025less} rely on pairwise comparisons between samples, such as similarity computation, clustering, or influence. While these local criteria encourage selecting samples that are different from one another, they do not explicitly account for the global structure of the dataset. Selecting samples that are locally diverse does not necessarily ensure that the overall representation space is well preserved. Furthermore, pairwise comparisons inevitably incur at least quadratic time complexity, posing a significant scalability bottleneck for large datasets.

In this work, we propose a fundamentally different perspective on multimodal data selection by adopting a global, subspace-oriented view. Instead of reasoning about relationships between individual samples, we focus on the contribution of each sample to the representation space of the entire dataset, i.e., the space spanned by the full dataset
representations.
This perspective builds on the observation that, under large-scale VIT, the induced representation space is often highly redundant and approximately low-rank, with most meaningful variability concentrated in a small number of dominant directions.
Accordingly, we aim to select a subset of samples whose representations can best span and approximate the dominant subspace of the original representation space. From this perspective, data selection can be viewed as a form of subspace compression in representation space: the goal is to identify a compact subset that preserves the dataset's essential global structure under a low-rank view, such that training on the subset remains close to training on the full data. This global formulation eliminates the need for pairwise comparisons and enables scalable data selection.

Based on this insight, we introduce ScalSelect, a scalable and training-free multimodal data selection method. ScalSelect constructs sample representations directly from the target model by leveraging instruction-conditioned attention signals in the first transformer layer of the LLM, capturing instruction-relevant visual information while suppressing irrelevant content.
To select informative samples, ScalSelect estimates the dominant representation subspace and scores samples by their contributions to this subspace, selecting a compact subset with linear-time scalability with respect to the number of samples and without proxy models or datasets.
Extensive experiments across multiple VLM architectures, datasets, selection budgets, and evaluation benchmarks show that ScalSelect can match full-data training closely using only a small fraction of the data.
Under a 16\% selection budget, ScalSelect achieves over 97.50\% of the performance of full-data training on LLaVA-Vicuna-7B across two datasets, and even surpasses full-data training on Qwen3-VL models in our experiments.
Our contributions can be summarized as follows:
\begin{itemize}
    \item Instruction-Conditioned Early Representation is introduced to construct sample representations by extracting instruction-relevant visual features from the first LLM transformer layer of the target VLM, enabling instruction-aware modeling.
    \item Subspace-Aware Global Selection is proposed as a global data selection perspective that prioritizes preserving the dominant representation subspace rather than relying on local, sample-to-sample relationships.
    \item ScalSelect is developed as a scalable training-free selection method with linear-time complexity, avoiding pairwise comparisons, proxy models, and additional training procedures.
    \item Extensive empirical studies, including cross-model and cross-dataset evaluations, selection budget scaling, ablation studies, and representation-level analyses, systematically validate the effectiveness, robustness, and key design choices of ScalSelect.
\end{itemize}
\section{Related Work}
\subsection{Vision-Language Models and Visual Instruction Tuning.}
Vision-language models (VLMs) have made significant progress by combining pretrained large language models with visual encoders, enabling models to perform various visual-language tasks. Early works, such as CLIP \cite{radford2021learning} and ALIGN \cite{jia2021scaling}, pioneered large-scale contrastive learning to align visual features with textual representations in a shared latent space.
Recently, visual instruction tuning (VIT) \cite{liu2023visual,liu2024improved,zhao2023svit} has advanced VLMs by fine-tuning them with large-scale image-instruction pairs, enabling models to follow natural language instructions for visual inputs. 
Pioneering models like LLaVA \cite{liu2023visual}, LLaVA-1.5 \cite{liu2024improved}, and MiniGPT-4 \cite{zhu2023minigpt} established the feasibility of following complex human intentions by VIT. This trajectory has been further advanced by diverse architectures, including Qwen3-VL \cite{bai2025qwen3vltechnicalreport}, LLaVA-NeXT \cite{liu2024llavanext}, and DeepSeek-VL2 \cite{wu2024deepseek}. Moreover, specialized models have pushed the boundaries in niche domains \cite{chen2024expanding,chen2024sharegpt4v,lin2025physbrain}, such as DeepSeek-OCR \cite{wei2025deepseek} for document intelligence and LLaVA-Med \cite{li2023llava} for biomedicine.
However, VIT faces challenges in terms of computational efficiency, as full fine-tuning on large datasets is expensive, and many samples in these datasets contribute redundantly to performance. This has led to the exploration of multimodal data selection to reduce computational costs while largely preserving model performance.

\subsection{Multimodal Data Selection for Visual Instruction Tuning.}
Multimodal data selection aims to identify informative subsets of image-instruction pairs for visual instruction tuning (VIT), reducing training cost while retaining most of the performance of full fine-tuning. Early approaches primarily rely on simple heuristics, such as random sampling, instruction length, or perplexity-based filtering \cite{marion2023less}, which are computationally cheap but largely task agnostic. More recent work has explored training-based \cite{liu2025less,wu2024icons,ma2025mllm,chen2024your} and gradient-based \cite{yang2024clip,maharana2024adapt} selection methods, which estimate sample importance via warm-up training, proxy models, or gradient backpropagation through the target model. While effective in some settings, these approaches introduce additional training stages, incur substantial computational overhead, and suffer from limited scalability on large multimodal datasets. To overcome these issues, a growing line of research focuses on training-free data selection methods \cite{lee2024concept,bi2025prism,ivison2025large,tang2025training,liu2025picking,yu2025mastering} that operate on pretrained representations without explicit training or backpropagation. 
However, existing training-free methods rely on external proxy encoders, construct instruction-agnostic sample representations, or depend on pairwise similarity computations, leading to representation mismatch, limited instruction awareness, and quadratic complexity.
For example, COINCIDE \cite{lee2024concept} selects samples based on clustering and similarity but relies on proxy encoders and pairwise comparisons. PRISM \cite{bi2025prism} depends on similarity computations with quadratic complexity. RDS+ \cite{ivison2025large} achieves linear-time selection but assumes access to an auxiliary proxy dataset that matches the evaluation distribution, which may not hold in practice.
These limitations highlight the need for more scalable and representation-faithful training-free data selection strategies for visual instruction tuning.

\begin{figure*}[!tbhp]
  \centering
  \includegraphics[width=1\linewidth]{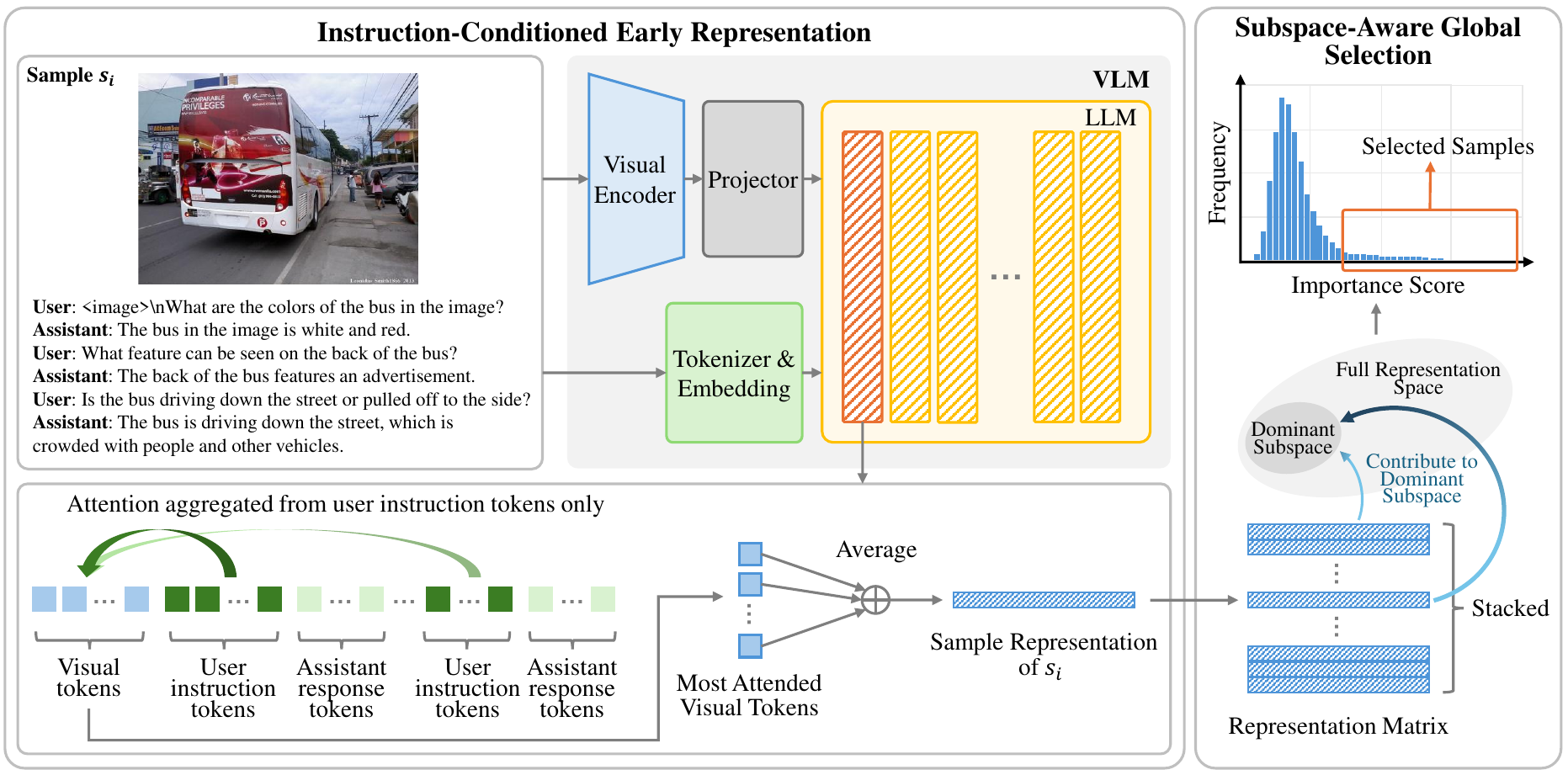}
   \caption{Overview of ScalSelect. (Left) For each sample, the target VLM extracts instruction-conditioned early representations by aggregating the visual tokens that are most attended by the user instruction tokens in the first layer of the LLM, yielding a compact sample representation. (Right) The representations of all samples are stacked into a representation matrix, from which ScalSelect identifies the dominant low-rank subspace of the full representation space (the space spanned by the full dataset representations). Each sample is scored according to its contribution to this dominant subspace, producing an importance score distribution from which a compact subset of informative samples is selected.}
   \label{fig:method}
\end{figure*}

\section{ScalSelect}\label{sec:method}
In this section, we describe ScalSelect, a training-free multimodal data selection method for visual instruction tuning.
Given a target VLM and a pool of visual instruction tuning data, ScalSelect proceeds in two stages.
First, it constructs sample representations via Instruction-Conditioned Early Representation based on the target model.
Second, based on the resulting sample representations, it performs Subspace-Aware Global Selection to identify a compact subset whose representations approximate the dominant subspace of the full dataset.
We next detail each component of ScalSelect.


\subsection{Instruction-Conditioned Early Representation}\label{sec:repr}

A growing body of representation learning related work \cite{feng2025align,zhang2025cross,lin2025boosting,sun2025transformer} suggests that the early transformer layers of the LLM play a critical role in cross-modal alignment.
For example, Align-KD \cite{feng2025align} empirically shows that the first LLM layer primarily aligns text and visual hidden states into a shared representation space, while the final layer projects representations into the output space.
Other works \cite{zhang2025cross,lin2025boosting} reveal that, in the early LLM layers, visual information is predominantly injected into textual representations, and visual tokens receive the highest attention weights, after which their attention rapidly diminishes to negligible levels.
Taken together, these empirical observations suggest a consistent layer-wise tendency in VLMs: \textit{early LLM layers tend to extract instruction-relevant visual information, the middle layers are more involved in text-dominant reasoning, and the final layer primarily maps the fused representation to the output space.}

In addition, for the same image, different instructions can induce the model to attend to entirely different visual regions and semantic attributes.
Consequently, treating visual representations independently of the instruction, as adopted by many existing data selection methods, fails to account for the diversity introduced by varying textual instructions.
This motivates us to condition sample representations on the input instruction when characterizing their contribution to the global representation space.

Motivated by these findings, we construct our sample representations from the first transformer layer of the LLM, where cross-modal alignment is most prominent, visual tokens receive the highest attention, and visual representations are already conditioned on the input instruction.
Specifically, let the original dataset to be selected be denoted as $D=\{ s_1, \ldots, s_N \}$, where each sample $s_i=\{V_i,U_i^1,A_i^1,\ldots,U_i^{T_i},A_i^{T_i}\}$
consists of the visual input $V_i$ and the $T_i$-turn conversation. Each conversation round $\{U_i^{t},A_i^{t}\}$ corresponds to the user instruction $U_i^{t}$ and the assistant response $A_i^{t}$ at turn $t \in \{1,\ldots,T_i\}$.
For each sample $s_i\in D$, the visual input is encoded and projected into the LLM hidden space, while the text input is tokenized and embedded into the same space:

\begin{equation}
\begin{aligned}
    &\{v_i^1,\ldots,v_i^{N_v}\} \leftarrow \operatorname{Proj}(\operatorname{VE}(V_i)), \\
    &\{u_i^{t,1},\ldots,u_i^{t,{N_u^t}}\} \leftarrow \operatorname{Emb}(\operatorname{Tok}(U_i^t)), \\
    &\{a_i^{t,1},\ldots,a_i^{t,{N_a^t}}\} \leftarrow \operatorname{Emb}(\operatorname{Tok}(A_i^t)),
\end{aligned}
\label{eq:encode}
\end{equation}
where $\operatorname{VE}(\cdot)$ denotes the visual encoder, $\operatorname{Proj}(\cdot)$ denotes the projector, $\operatorname{Tok}(\cdot)$ and $\operatorname{Emb}(\cdot)$ denote the tokenizer and embedding layers of the LLM, respectively.
$\{v_i^j\}_{j=1}^{N_v}$ denotes the sequence of visual token embeddings, while $\{u_i^{t,j}\}_{j=1}^{N_u^t}$ and $\{a_i^{t,j}\}_{j=1}^{N_a^t}$ denote the token embeddings of the user instruction and assistant response at turn $t$, respectively.

At the first transformer layer of the LLM, multi-head self-attention is applied over the concatenated sequence of visual and textual tokens.
Let $Q^{(1)}, K^{(1)}\in \mathbb{R}^{L \times d_h}$ denote the query and key matrices at this layer, where $L$ is the total number of tokens in the input sequence and $d_h$ is the hidden dimension of each attention head.
For a single attention head, the attention matrix is computed as
\begin{equation}
\mathcal{A}^{(1)} = \operatorname{softmax}\!\left(\frac{Q^{(1)} {K^{(1)}}^\top}{\sqrt{d_h}}\right),
\end{equation}
where $\mathcal{A}^{(1)} \in \mathbb{R}^{L \times L}$ encodes post-softmax attention scores between all token pairs, and the entry $\mathcal{A}^{(1)}[p,q]$ represents the attention score from the $p$-th token to the $q$-th token in the input sequence.
In the multi-head setting, we average the attention matrices across all heads.
Based on the head-averaged attention matrix, we compute the aggregated attention score received by each visual token $\alpha_i^{j}$ ($j\in \{1,\ldots, N_v\}$) from all user instruction tokens:

\begin{equation}
\begin{aligned}
    \alpha_i^{j} = \sum_{t=1}^{T_i} \sum_{m=1}^{N_u^t} 
\operatorname{Attn}^{(1)}(u_i^{t,m}, v_i^{j}),
\end{aligned}
\label{eq:atten}
\end{equation}
where $\operatorname{Attn}^{(1)}(\cdot,\cdot)$ denotes the attention score from a user instruction token to a visual token in the head-averaged attention matrix of the first transformer layer.

We then sort the visual tokens in descending order of their aggregated attention scores and select tokens from top to bottom until their cumulative attention reaches a threshold $\tau \in (0,1)$ of the total attention.
For each sample $s_i\in D$, the selected visual tokens form the set of the most attended visual tokens, denoted as $\mathcal{V}_i$:

\begin{equation}
    \mathcal{V}_i
=
{\arg\min}_{\mathcal{V}_i' \subseteq \{v_i^1,\ldots,v_i^{N_v}\}}
|\mathcal{V}_i'|
\quad
\text{s.t.}
\
\sum_{v_i^{j} \in \mathcal{V}_i'} \alpha_i^{j}
\ge
\tau \cdot \sum_{j=1}^{N_v} \alpha_i^{j}.
\label{eq:topk}
\end{equation}
We find that the selection remains stable when the attention-mass threshold $\tau$ varies within a reasonable range (e.g., 0.85--0.95; see Appendix \ref{sec:threshold}). We therefore use $\tau=0.9$ by default for a good trade-off between compactness and coverage.

Finally, we average the hidden states of the selected visual tokens at the first transformer layer to obtain the representation of sample $s_i$:

\begin{equation}
\mathbf{x}_i = \frac{1}{|\mathcal{V}_i|}
\sum_{v_i^{j} \in \mathcal{V}_i} \operatorname{LLM}^{(1)}(v_i^{j}),
\label{eq:pool}
\end{equation}
where $\operatorname{LLM}^{(1)}(\cdot)$ denotes the output hidden state of the first transformer layer.

\subsection{Subspace-Aware Global Selection}

Most existing data selection methods adopt a local view by explicitly modeling relationships between samples, typically through pairwise similarity or influence-based measures, and make selection decisions based on local, sample-to-sample comparisons.
In contrast, we move beyond modeling pairwise relationships between samples and instead focus on how each individual sample relates to the global representation space as a whole.
Building on this view, we take a global, subspace-aware perspective: \textit{we aim to identify a subset of representations that best span and approximate the subspace of the original representation space}, where \textit{subspace} refers to the dominant low-rank principal subspace of the sample representation matrix, i.e., the linear space spanned by the top singular directions that capture the majority of variance in the data.
The selected subset retains most of the information relevant to the best rank-$k$ approximation of the full dataset, and is thus nearly as informative as the original data under a low-rank representation view.
By focusing on this dominant subspace, the selected subset is expected to better capture the global representational characteristics of the full dataset, which empirically helps models trained on it achieve performance comparable to training on the full dataset.

To this end, we adopt a leverage-score-based row selection strategy inspired by CUR matrix decompositions \cite{drineas2008relative,mahoney2009cur}.
CUR decomposition approximates a matrix by selecting a subset of its rows and columns such that the selected rows and columns capture the dominant low-rank structure of the original matrix.
Given the attention-conditioned sample representations $\{\mathbf{x}_1,\ldots,\mathbf{x}_N\}$ obtained in Section \ref{sec:repr}, we stack all sample representations into a matrix:
\begin{equation}
    \mathbf{X}=[\mathbf{x}_1;\mathbf{x}_2;\ldots;\mathbf{x}_N]\in \mathbb{R}^{N \times d},
\label{eq:matrix}
\end{equation}
where each row of $\mathbf{X}$ is a sample representation.
We first perform column-wise centering on $\mathbf{X}$ to obtain $\mathbf{X}_c$.

We then estimate the dominant low-rank structure of $\mathbf{X}_c$ via truncated singular value decomposition (SVD) \cite{wall2003singular},
focusing only on the leading singular components:
\begin{equation}
    \mathbf{X}_c \;\approx\; \mathbf{U}_k \mathbf{\Sigma}_k \mathbf{V}_k^\top,
\end{equation}
where $\mathbf{\Sigma}_k=\text{diag} (\sigma_1 , \ldots , \sigma_k)\in \mathbb{R}^{k \times k}$,
and $\mathbf{U}_k=[\boldsymbol{s}_1,\ldots,\boldsymbol{s}_k]\in \mathbb{R}^{N \times k}$.
Here, $\{ \sigma_1 , \ldots , \sigma_k\}$ ($ \sigma_1 \geq \sigma_2 \geq \ldots \geq \sigma_k \geq 0$)
denote the top-$k$ singular values of $\mathbf{X}_c$, and
$\{\boldsymbol{s}_1,\ldots,\boldsymbol{s}_k\}$ are the corresponding left singular vectors.

For truncated SVD, we select the smallest $k$ such that the cumulative energy of the top-$k$
singular values reach 90\% of the total spectral energy
(following common practice \cite{mahoney2009cur,abdi2010principal,
eckart1936approximation,halko2011finding} in low-rank approximation, we set the energy threshold to $90\%$ throughout this work):
\begin{equation}
k=
{\arg\min}_{k' \in \{1,\ldots,r\}} k'
\quad \text{s.t.} \
\sum_{j=1}^{k'} \sigma_j^2
\;\ge\;
90\% \cdot \sum_{j=1}^{r} \sigma_j^2,
\label{eq:energy}
\end{equation}
where $r=\operatorname{rank}(\mathbf{X}_c)$ is the rank of $\mathbf{X}_c$, and $k$ denotes the dimensionality of the selected low-rank subspace, corresponding to the minimum number of dominant singular components required to explain 90\% of the total spectral energy.



After that, we calculate an ``importance score'' (statistical leverage score) for each row of $\mathbf{X}_c$ (i.e., each sample):
\begin{equation}
 \pi_i=\sum_{j=1}^k\left(\boldsymbol{s}_j^{i}\right)^2,
\label{eq:score}
\end{equation}
where $i\in \{1,\ldots,N\}$, and $\boldsymbol{s}^i_j$ is the $i$-th element of the $j$-th left singular vector of $\mathbf{X}_c$. 
The importance score has a natural meaning: the ``statistical leverage'' or ``influence'' of the row on the best low-rank fit of matrix $\mathbf{X}_c$.
Finally, after obtaining the importance score for each sample, we select the top-ranked samples according to their importance scores to form the final subset.

Overall, by prioritizing samples with the highest statistical leverage, the subset preserves most of the information relevant to reconstructing the best rank-$k$ approximation of the original representation matrix.
In this sense, our method achieves effective representational compression, retaining the dominant structural information of the original data.

\section{Scalability Analysis}
In this section, we analyze the scalability of ScalSelect with respect to the number of samples $N$, and compare it with other multimodal data selection methods to demonstrate its scalability advantage.

\textbf{Scalability of ScalSelect.}
In the Attention-Conditioned Early Representation stage, each sample requires only a single forward pass, and the method relies solely on the attention scores and hidden states from the first transformer layer of the LLM. Therefore, the time complexity of this stage is $O(N)$.
In the Subspace-Aware Global Selection stage, the overall time complexity mainly consists of the following components:
1) Performing truncated SVD on the representation matrix $\mathbf{X}_c$, which computes only the top-$k$ singular components $\mathbf{U}_k \in \mathbb{R}^{N \times k}$ and
$\mathbf{\Sigma}_k \in \mathbb{R}^{k \times k}$.
This step costs $O(N d k)$ time.
2) Computing the importance scores, which has a complexity of $O(Nk)$.
Since the representation dimension $d \ll N$ is fixed by the model architecture and the subspace dimensionality $k \ll d \ll N$ is consistently small in practice (e.g., $k=9$ for LLaVA-V-625K as presented in Section \ref{sec:ablation}), the overall selection process scales linearly with the dataset size $N$.

\textbf{Scalability Comparison with Other Methods.}
Existing multimodal data selection methods based on training or gradients depend on proxy training or gradient backpropagation, introducing additional training and considerable computational overhead.
Among training-free methods, several approaches depend on computing pairwise similarities or influence scores between samples, resulting in quadratic time complexity $O(N^2)$ and limited scalability.
Among recent training-free methods, COINCIDE \cite{lee2024concept} has a time complexity of $O\!\left(KN + \sum \lvert C_i \rvert^2\right)$, when the sample number in each cluster $C_i$ is balanced, its complexity reduces to $O(KN+N^2/K)$; however, it still relies on a proxy model.
PRISM \cite{bi2025prism} computes pairwise similarities between samples, leading to a quadratic time complexity of $O(N^2)$.
RDS+ \cite{ivison2025large} achieves linear-time complexity, but relies on a strong assumption that a proxy dataset sharing the same data distribution as the evaluation set is available, which is unrealistic in practice and may introduce distribution bias.
Overall, compared with these methods, ScalSelect achieves linear-time complexity in a training-free manner, without relying on any external models or proxy datasets, demonstrating clear advantages in scalability.

\section{Experiments}
\subsection{Experiment Setup}
\textbf{Dataset.}
We conduct experiments on a large-scale multimodal visual instruction tuning dataset, LLaVA-V-625K, which we construct from LLaVA-665K \cite{liu2024improved} by removing 40K text-only samples and retaining the remaining 625K multimodal samples.
LLaVA-V-625K comprises diverse multimodal instruction data, covering tasks such as visual question answering (VQA), optical character recognition (OCR), region-level VQA, and visual conversation.
To verify the generalization of our method across different datasets, we additionally evaluate on the public 180K GPT-4-generated dataset of the LRV-Instruction \cite{liu2023mitigating}, which we refer to as LRV-Sub-180K. This dataset contains diverse visual question answering data with varied instruction styles and visual content.
Unless otherwise specified, all experiments are conducted on LLaVA-V-625K with a selection budget of 100K samples (16\%).

\textbf{Models.}
Following the training setup of LLaVA-1.5 \cite{liu2024improved}, we use the pretrained model before visual instruction tuning on LLaVA-1.5-7B, which we refer to as LLaVA-Vicuna-7B, as the target model for most of our experiments.
To verify the generality of our method across different model architectures, we additionally evaluate on two widely adopted VLMs, Qwen3-VL-4B-Instruct and Qwen3-VL-8B-Instruct \cite{bai2025qwen3vltechnicalreport}.
All models are fine-tuned for one epoch on 8 $\times$ NVIDIA H100 (80GB) using the same training configuration, ensuring fair and controlled comparisons.
Training hyperparameters are provided in Appendix \ref{sec:details}.

\textbf{Evaluation Benchmarks.}
We evaluate the fine-tuned models on a diverse set of widely used multimodal benchmarks to assess various capabilities of VLMs.
To evaluate the overall capabilities of multimodal large language models, including multimodal understanding, coarse-grained and fine-grained perception, as well as cognition and reasoning, we adopt three widely used benchmarks: 1) MMBench \cite{liu2024mmbench}, which includes an English version (MMBench-En) and a Chinese version (MMBench-Cn); 2) MME \cite{fu2025mme}, which comprises a perception subset (MME-P) and a cognition subset (MME-C); and 3) AI2D \cite{kembhavi2016diagram}, a benchmark for diagram understanding.
To assess hallucination tendencies, we adopt POPE \cite{li2023evaluating}, reporting its two core metrics: accuracy (POPE-A) and precision (POPE-P).
For scientific question answering, we adopt ScienceQA-IMG (SQA-IMG), the vision-based subset of ScienceQA \cite{lu2022learn}, for evaluation.
For OCR tasks, we adopt OCRBench \cite{liu2024ocrbench} for evaluation.

\textbf{Baselines.}
We compare ScalSelect with a range of representative and widely used multimodal data selection methods. 
Traditional heuristics include Random, Length, and Perplexity \cite{marion2023less}, where Random randomly selects samples, Length uses sample length as an importance score, and Perplexity selects samples with medium perplexity values.
For strong recent representative methods, we compare COINCIDE \cite{lee2024concept}, PRISM \cite{bi2025prism}, and RDS+ \cite{ivison2025large}, which are widely adopted and commonly reported to achieve competitive performance under various experimental settings in prior work.
We focus on baselines applicable without additional optimization or warm-up training for fair comparison.
We also report Full-Finetune results using the entire training dataset as an upper-bound reference.
``Rel.'' denotes the average relative performance across benchmarks, computed as the mean percentage normalized by the reference method set to 100\% in each table.

\begin{table*}[!thp]
\centering
\renewcommand{\arraystretch}{1.2}
\setlength\tabcolsep{8pt} 
\resizebox{\linewidth}{!}
{
\begin{tabular}{l|ccccccccc|c}
\toprule
\textbf{Method} & \textbf{MMBench-En} & \textbf{MMBench-Cn} & \textbf{MME-P} & \textbf{MME-C} & \textbf{AI2D} & \textbf{POPE-A} & \textbf{POPE-P} & \textbf{SQA-IMG} & \textbf{OCRBench} & \textbf{Rel.}\\
\midrule
Full-Finetune & 63.96 & 56.73 & 1463.87 & 278.57 & 53.79 & 85.24 & 93.05 & 67.63 & 20.30 & 100.00\% \\
\midrule
Random & 57.96 & 51.74 & 1418.85 & 295.36 & 50.78 & 84.96 & 90.71 & 65.20 & 17.90 & 95.66\% \\
Length & 49.10 & 36.43 & 1282.97 & 292.14 & 37.44 & 81.82 & 96.71 & 52.70 & 13.60 & 83.10\% \\
Perplexity & 31.74 & 32.47 & 1403.08 & 278.93 & 38.02 & 84.81 & 93.02 & 48.44 & 18.60 & 81.80\% \\
COINCIDE & 62.16 & 55.44 & 1422.01 & 274.29 & 51.70 & 83.79 & 90.99 & 66.83 & 18.30 & 96.85\% \\
PRISM & 61.49 & 52.96 & 1396.24 & 276.43 & 50.87 & 83.38 & 93.26 & 64.06 & 18.80 & 96.01\% \\
RDS+ & 56.05 & 50.16 & 1414.66 & 302.85 & 50.84 & 84.84 & 92.15 & 64.45 & 18.60 & 95.71\% \\
\rowcolor{navyblue!10}ScalSelect (Ours) & 59.19 & 52.80 & 1400.34 & 308.57 & 51.75 & 83.96 & 94.73 & 65.29 & 19.40 & 97.85\% \\
\bottomrule
\end{tabular}
}
\caption{Performance comparison with baselines on LLaVA-V-625K under a 100K selection budget.}
\label{tab:exp_main}
\end{table*}

\begin{table*}[!thp]
\centering
\renewcommand{\arraystretch}{1.2}
\setlength\tabcolsep{8pt} 
\resizebox{\linewidth}{!}
{
\begin{tabular}{l|l|ccccccccc|c}
\toprule
\textbf{Method} & \textbf{\#Samples} & \textbf{MMBench-En} & \textbf{MMBench-Cn} & \textbf{MME-P} & \textbf{MME-C} & \textbf{AI2D} & \textbf{POPE-A} & \textbf{POPE-P} & \textbf{SQA-IMG} & \textbf{OCRBench} & \textbf{Rel.}\\
\midrule
Full-Finetune & 625K & 63.96 & 56.73 & 1463.87 & 278.57 & 53.79 & 85.24 & 93.05 & 67.63 & 20.30 & 100.00\% \\
\midrule
\multirow{5}{*}{ScalSelect} & 50K & 57.79 & 51.68 & 1282.22 & {314.29} & 50.87 & 83.99 & {93.64} & 63.11 & 18.10 & 95.34\% \\
& 100K & 59.19 & 52.80 & 1400.34 & {308.57} & 51.75 & 83.96 & {94.73} & 65.29 & 19.40 & 97.85\% \\
& 200K & 61.10 & 54.15 & 1435.79 & {298.93} & 53.21 & 84.50 & {93.87} & 65.39 & 19.50 & 98.67\% \\
& 300K & 63.05 & 55.09 & {1517.26} & {290.00} & {54.18} & 84.70 & {93.05} & 66.53 & 19.30 & 99.66\% \\
& 400K & {65.30} & {58.97} & 1425.41 & {299.29} & {54.21} & {85.74} & {93.60} & {67.72} & 19.80 & {101.16\%} \\
\bottomrule
\end{tabular}
}
\caption{Performance under different selection budgets. ``\#Samples'' denotes the number of samples selected from LLaVA-V-625K for fine-tuning.}
\label{tab:exp_samplenum}
\end{table*}

\subsection{Main Results}\label{sec:main_res}
\textbf{Performance Compared to Baselines.}
Table \ref{tab:exp_main} compares ScalSelect with a range of representative baselines on LLaVA-V-625K under a fixed 100K selection budget.
We observe that simple heuristic methods, including Length and Perplexity, consistently underperform Random sampling across most benchmarks, indicating that instruction length or perplexity alone is insufficient for identifying important training samples, and tend to induce skewed and less informative sample compositions.
COINCIDE exhibits mixed performance across benchmarks, achieving competitive results on some evaluation tasks (e.g., MMBench) while showing clear weaknesses on MME-C and POPE-P. 
This pattern indicates that emphasizing instance-level diversity can benefit MMBench, but may come at the cost of degraded reasoning robustness.
On another broad-coverage benchmark, MME, ScalSelect significantly outperforms COINCIDE, achieving a substantially higher score on MME-C while maintaining competitive performance on MME-P.
These results reveal a trade-off between performance on different broad-coverage benchmarks, such as MMBench and MME, reflecting different emphases on model capability.
Furthermore, ScalSelect consistently outperforms all baselines on MME-C, AI2D, and OCRBench, and matches or slightly surpasses Full-Finetune on MME-C and POPE-P. 
ScalSelect also achieves a clear advantage on OCRBench, demonstrating its effectiveness on tasks requiring fine-grained, instruction-relevant visual attention (see Section \ref{sec:ablation} for further analysis).
Overall, under the 100K selection budget, ScalSelect achieves the highest average relative performance among all evaluated methods, reaching 97.85\% of Full-Finetune performance. These results suggest that ScalSelect offers a favorable balance between scalability and effectiveness under limited data budgets.

\textbf{Performance Under Different Selection Budgets.}
Table \ref{tab:exp_samplenum} reports the performance of ScalSelect on LLaVA-V-625K under different selection budgets. As the number of selected samples increases from 50K to 400K, the performance consistently improves across most benchmarks, demonstrating the effectiveness and stability of ScalSelect under varying selection budgets.
With only 50K selected samples, ScalSelect already retains over 95\% of the Full-Finetune performance on average.
Notably, when selecting 300K samples (less than half of the full dataset), ScalSelect attains performance that is nearly on par with Full-Finetune, achieving 99.66\% relative performance.
With 400K selected samples, ScalSelect surpasses Full-Finetune with a relative performance of 101.16\%, and achieves comparable or better results on most benchmarks, demonstrating that effective data selection can mitigate redundancy and improve training efficiency.
In addition, we observe an apparent trade-off between MME-P and MME-C. Specifically, as the number of selected samples grows from 50K to 300K, performance on MME-P consistently improves, while performance on MME-C shows a mild decline. At 400K, the two metrics become more balanced. This observation indicates that different evaluation aspects may respond differently to the selection budget, highlighting the importance of balancing perceptual and cognitive performance.

\begin{table*}[!thp]
\centering
\renewcommand{\arraystretch}{1.2}
\setlength\tabcolsep{8pt} 
\resizebox{\linewidth}{!}
{
\begin{tabular}{l|l|l|ccccccccc|c}
\toprule
\textbf{Model} & \textbf{Method} & \textbf{\#Samples} & \textbf{MMBench-En} & \textbf{MMBench-Cn} & \textbf{MME-P} & \textbf{MME-C} & \textbf{AI2D} & \textbf{POPE-A} & \textbf{POPE-P} & \textbf{SQA-IMG} & \textbf{OCRBench} & \textbf{Rel.}\\
\midrule
\multirow{2}{*}{LLaVA-Vicuna-7B} & Full-Finetune & 625K & 63.96 & 56.73 & 1463.87 & 278.57 & 53.79 & 85.24 & 93.05 & 67.63 & 20.30 & 100.00\% \\
 & \cellcolor{navyblue!10}ScalSelect & \cellcolor{navyblue!10}100K & \cellcolor{navyblue!10}59.19 & \cellcolor{navyblue!10}52.80 & \cellcolor{navyblue!10}1400.34 & \cellcolor{navyblue!10}308.57 & \cellcolor{navyblue!10}51.75 & \cellcolor{navyblue!10}83.96 & \cellcolor{navyblue!10}94.73 & \cellcolor{navyblue!10}65.29 & \cellcolor{navyblue!10}19.40 & \cellcolor{navyblue!10}97.85\% \\
\midrule
\multirow{2}{*}{Qwen3-VL-4B-Instruct} & Full-Finetune & 625K & 82.06 & 71.47 & 1589.11 & 651.43 & 78.95 & 88.37 & 97.03 & 87.26 & 69.10 & 100.00\% \\
 & \cellcolor{navyblue!10}ScalSelect & \cellcolor{navyblue!10}100K & \cellcolor{navyblue!10}84.30 & \cellcolor{navyblue!10}81.39 & \cellcolor{navyblue!10}1652.75 & \cellcolor{navyblue!10}658.93 & \cellcolor{navyblue!10}79.92 & \cellcolor{navyblue!10}88.17 & \cellcolor{navyblue!10}96.86 & \cellcolor{navyblue!10}91.03 & \cellcolor{navyblue!10}75.40 & \cellcolor{navyblue!10}104.00\% \\
\midrule
\multirow{2}{*}{Qwen3-VL-8B-Instruct} &Full-Finetune & 625K & 84.30 & 82.74 & 1700.21 & 676.43 & 80.47 & 88.54 & 96.39 & 88.40 & 66.00 & 100.00\% \\
 & \cellcolor{navyblue!10}ScalSelect & \cellcolor{navyblue!10}100K & \cellcolor{navyblue!10}82.46 & \cellcolor{navyblue!10}82.06 & \cellcolor{navyblue!10}1709.07 & \cellcolor{navyblue!10}701.43 & \cellcolor{navyblue!10}81.83 & \cellcolor{navyblue!10}87.97 & \cellcolor{navyblue!10}94.39 & \cellcolor{navyblue!10}92.12 & \cellcolor{navyblue!10}77.30 & \cellcolor{navyblue!10}102.39\% \\
\bottomrule
\end{tabular}
}
\caption{Performance across different models. Results are reported on LLaVA-V-625K with LLaVA-Vicuna-7B, Qwen3-VL-4B-Instruct, and Qwen3-VL-8B-Instruct under the same selection budget of 100K samples.}
\label{tab:exp_acrossmodel}
\end{table*}

\begin{table*}[!thp]
\centering
\renewcommand{\arraystretch}{1.2}
\setlength\tabcolsep{8pt} 
\resizebox{\linewidth}{!}
{
\begin{tabular}{l|l|l|ccccccccc|c}
\toprule
\textbf{Dataset} & \textbf{Method} & \textbf{\#Samples} & \textbf{MMBench-En} & \textbf{MMBench-Cn} & \textbf{MME-P} & \textbf{MME-C} & \textbf{AI2D} & \textbf{POPE-A} & \textbf{POPE-P} & \textbf{SQA-IMG} & \textbf{OCRBench} & \textbf{Rel.}\\
\midrule
\multirow{3}{*}{LLaVA-V-625K} & Full-Finetune & 625K & 63.96 & 56.73 & 1463.87 & 278.57 & 53.79 & 85.24 & 93.05 & 67.63 & 20.30 & 100.00\% \\
& Random & 100K & 57.96 & 51.74 & 1418.85 & 295.36 & 50.78 & 84.96 & 90.71 & 65.20 & 17.90 & 95.66\% \\
 & \cellcolor{navyblue!10}ScalSelect & \cellcolor{navyblue!10}100K & \cellcolor{navyblue!10}59.19 & \cellcolor{navyblue!10}52.80 & \cellcolor{navyblue!10}1400.34 & \cellcolor{navyblue!10}308.57 & \cellcolor{navyblue!10}51.75 & \cellcolor{navyblue!10}83.96 & \cellcolor{navyblue!10}94.73 & \cellcolor{navyblue!10}65.29 & \cellcolor{navyblue!10}19.40 & \cellcolor{navyblue!10}97.85\% \\
\midrule
\multirow{3}{*}{LRV-Sub-180K} & Full-Finetune & 180K & 34.19 & 22.20 & 1082.21 & 258.93 & 40.97 & 75.36 & 69.13 & 58.60 & 16.50 & 100.00\% \\
& Random & 29K & 35.98 & 20.75 & 848.76 & 233.93 & 41.32 & 60.18 & 62.37 & 58.06 & 16.50 & 93.05\% \\
 & \cellcolor{navyblue!10}ScalSelect & \cellcolor{navyblue!10}29K & \cellcolor{navyblue!10}36.30 & \cellcolor{navyblue!10}21.69 & \cellcolor{navyblue!10}840.59 & \cellcolor{navyblue!10}281.79 & \cellcolor{navyblue!10}43.39 & \cellcolor{navyblue!10}65.04 & \cellcolor{navyblue!10}62.69 & \cellcolor{navyblue!10}60.78 & \cellcolor{navyblue!10}16.60 & \cellcolor{navyblue!10}97.51\% \\
\bottomrule
\end{tabular}
}
\caption{Performance across different datasets. Results are reported with LLaVA-Vicuna-7B on LLaVA-V-625K and LRV-Sub-180K under the same selection budget of 16\%.}
\label{tab:exp_acrossdataset}
\end{table*}

\begin{table}[!thp]
\centering
\renewcommand{\arraystretch}{1.2}
\setlength\tabcolsep{8pt} 
\resizebox{\linewidth}{!}
{
\begin{tabular}{l|c|cccccccc}
\toprule
\textbf{Method} & \textbf{625K} & \textbf{100K} & \textbf{300K} & \textbf{400K} \\
\midrule
Train projector & 100.00\% & 97.85\% & 99.66\% & 101.16\%  \\
Freeze projector & 100.00\% & 98.30\% & 99.06\% & 100.20\% \\
\bottomrule
\end{tabular}
}
\caption{Performance under different projector freezing settings. Target model is LLaVA-Vicuna-7B. All settings freeze the visual encoder. ``Freeze Projector'' fine-tunes only the LLM, while ``Train Projector'' fine-tunes both the LLM and the projector.}
\label{tab:exp_projector}
\end{table}

\subsection{Extended Results}
\textbf{Performance Across Different Models.}
Table \ref{tab:exp_acrossmodel} presents the performance of ScalSelect with 100K selected samples on LLaVA-V-625K across different model architectures, including LLaVA-Vicuna-7B, Qwen3-VL-4B-Instruct, and Qwen3-VL-8B-Instruct.
Compared to LLaVA-Vicuna-7B, Qwen3-VL-4B-Instruct and Qwen3-VL-8B-Instruct exhibit stronger performance under the 100K selection budget, both slightly exceeding Full-Finetune.
This observation suggests that more capable models may be more sensitive to data redundancy and can benefit more from high-quality and informative subsets selected by ScalSelect.

\textbf{Performance Across Different Datasets.}
Table \ref{tab:exp_acrossdataset} presents the performance of ScalSelect with LLaVA-Vicuna-7B across different datasets, including LLaVA-V-625K and LRV-Sub-180K, under the same selection budget of 16\%.
On both datasets, ScalSelect consistently outperforms Random sampling and achieves performance close to Full-Finetune.
Notably, on LRV-Sub-180K, ScalSelect with only 29K selected samples attains 97.51\% of the Full-Finetune performance, surpassing Random sampling by a clear margin.
These results suggest that ScalSelect generalizes well across datasets with different instruction styles and data distributions.

\textbf{Performance Under Different Projector Freezing Settings.}
Table \ref{tab:exp_projector} compares ScalSelect under two commonly used visual instruction tuning paradigms: freezing the projector and fine-tuning only the LLM, or fine-tuning both the LLM and the projector. Overall, the performance differences between the two settings are relatively small across different selection budgets.
This indicates that ScalSelect is robust to different projector freezing strategies and can be effectively applied under both mainstream visual instruction tuning paradigms.

\begin{table*}[!thp]
\centering
\renewcommand{\arraystretch}{1.2}
\setlength\tabcolsep{8pt} 
\resizebox{\linewidth}{!}
{
\begin{tabular}{l|c|ccccccccc|c}
\toprule
\textbf{Method} & \textbf{k} & \textbf{MMBench-En} & \textbf{MMBench-Cn} & \textbf{MME-P} & \textbf{MME-C} & \textbf{AI2D} & \textbf{POPE-A} & \textbf{POPE-P} & \textbf{SQA-IMG} & \textbf{OCRBench} & \textbf{Rel.}\\
\midrule
\rowcolor{navyblue!10}ScalSelect (Ours) & 9 & 59.19 & 52.80 & 1400.34 & 308.57 & 51.75 & 83.96 & 94.73 & 65.29 & 19.40 & 100.00\% \\
ScalSelect-NoInsCon & 18 & 61.72 & 55.44 & 1394.02 & 288.57 & 50.84 & 84.82 & 92.24 & 64.40 & 17.90 & 98.88\% \\
ScalSelect-NoCenter & 1 & 55.21 & 48.04 & 1356.01 & 270.00 & 51.18 & 84.06 & 91.10 & 66.29 & 19.30 & 96.09\% \\
\bottomrule
\end{tabular}
}
\caption{Influence of instruction-conditioning \& Influence of column-wise centering. ``ScalSelect-NoInsCon'' removes instruction conditioning by selecting all visual tokens without conditioning on instruction-to-vision attention signals.
``ScalSelect-NoCenter'' disables column-wise centering before subspace decomposition.
``$k$'' denotes the selected subspace dimensionality as described in Section \ref{sec:method}.}
\label{tab:exp_ablation_1}
\end{table*}

\begin{table*}[!thp]
\centering
\renewcommand{\arraystretch}{1.2}
\setlength\tabcolsep{8pt} 
\resizebox{\linewidth}{!}
{
\begin{tabular}{l|c|ccccccccc|c}
\toprule
\textbf{Method} & \textbf{k} & \textbf{MMBench-En} & \textbf{MMBench-Cn} & \textbf{MME-P} & \textbf{MME-C} & \textbf{AI2D} & \textbf{POPE-A} & \textbf{POPE-P} & \textbf{SQA-IMG} & \textbf{OCRBench} & \textbf{Rel.}\\
\midrule
\rowcolor{navyblue!10}First Layer (Ours) & 9 & 59.19 & 52.80 & 1400.34 & 308.57 & 51.75 & 83.96 & 94.73 & 65.29 & 19.40 & 100.00\% \\
Middle Layer & 59 & 57.34 & 50.45 & 1413.59 & 260.00 & 48.87 & 82.11 & 94.43 & 63.06 & 18.70 & 95.84\% \\
Deep Layer & 260 & 59.53 & 52.86 & 1357.62 & 284.29 & 50.74 & 85.02 & 90.87 & 64.85 & 19.60 & 98.37\% \\
\bottomrule
\end{tabular}
}
\caption{Influence of LLM Layer Selection. ``$k$'' denotes the selected subspace dimensionality as described in Section \ref{sec:method}. ``First Layer'' (Ours) uses the hidden states from the first transformer layer, ``Middle Layer'' uses the hidden states from a middle transformer layer (the 16th layer), and ``Deep Layer'' uses the hidden states from the penultimate transformer layer.}
\label{tab:exp_ablation_2}
\end{table*}

\subsection{Ablation and Further Analysis}\label{sec:ablation}

\textbf{Influence of Instruction-Conditioning.}
We analyze the influence of instruction conditioning by comparing ScalSelect with ScalSelect-NoInsCon, where all visual tokens are selected without conditioning on instruction-to-vision attention signals. As shown in Table \ref{tab:exp_ablation_1}, removing instruction conditioning leads to an overall degradation in performance, suggesting that incorporating instruction-relevant attention cues is beneficial for constructing effective sample representations.
In particular, methods that incorporate instruction conditioning substantially outperform ScalSelect-NoInsCon on OCRBench, highlighting the importance of instruction-conditioned cues for tasks that require focusing on localized and instruction-relevant visual regions.
By selecting visual tokens that receive high attention from text instructions, ScalSelect helps emphasize instruction-relevant visual regions while reducing the influence of task-irrelevant areas.
This effect is further reflected in the learned subspace dimensionality $k$. ScalSelect-NoInsCon results in a larger $k$ value, which is consistent with more redundant information of its representation space.

\textbf{Influence of Column-Wise Centering.}
We further investigate the impact of column-wise centering by comparing ScalSelect with ScalSelect-NoCenter, where centering is not applied before subspace decomposition. As shown in Table \ref{tab:exp_ablation_1}, removing column-wise centering leads to a significant performance degradation, with the selected subspace dimensionality collapsing to $k=1$.
This behavior suggests that, without centering, the mean component of the representation matrix dominates the variance structure. As a result, the leading singular vector primarily captures the global mean of the representations, while meaningful variations among samples are largely suppressed. Consequently, the principal components tend to be ``absorbed'' by the mean, effectively flattening the representation space and diminishing sample diversity.
By performing column-wise centering, ScalSelect removes the global mean effect and allows the subspace decomposition to focus on variance-driven directions that reflect genuine differences among sample representations. This leads to a richer and more expressive subspace with higher effective dimensionality, which in turn supports more reliable importance estimation and improved selection performance.

\textbf{Influence of LLM Layer Selection.}
We study the effect of LLM layer selection by constructing sample representations from different transformer layers, including the first layer, a middle layer (the 16th layer), and the penultimate layer. For the deep setting, we use the penultimate layer instead of the final layer, as the last layer is typically more specialized toward output generation and may discard fine-grained representational information \cite{ethayarajh2019contextual}.
As shown in Table \ref{tab:exp_ablation_2}, using hidden states from the middle layer leads to the worst overall performance. This suggests that intermediate layers may place less emphasis on fine-grained visual grounding and instead focus more on abstract or mixed-modal representations, making them less suitable for sample representation in data selection.
Using hidden states from the deep layer improves performance on OCRBench, even surpassing the first-layer setting.
This observation suggests that later LLM layers may encode more detailed or layout-sensitive visual cues that are beneficial for OCR-related tasks, which is consistent with prior empirical findings \cite{xu2020layoutlm,xu2021layoutlmv2}.
Overall, extracting hidden states from the first transformer layer yields the best and most balanced performance across benchmarks.
We also observe that the selected subspace dimensionality $k$ increases substantially when using deeper LLM layers, indicating a more complex and less compressible representation space compared to early layers.
This result suggests that early LLM layers capture informative cross-modal interactions and provide a stable and effective basis for sample representation and data selection.

\begin{figure}[!tbhp]
  \centering
  \includegraphics[width=1\linewidth]{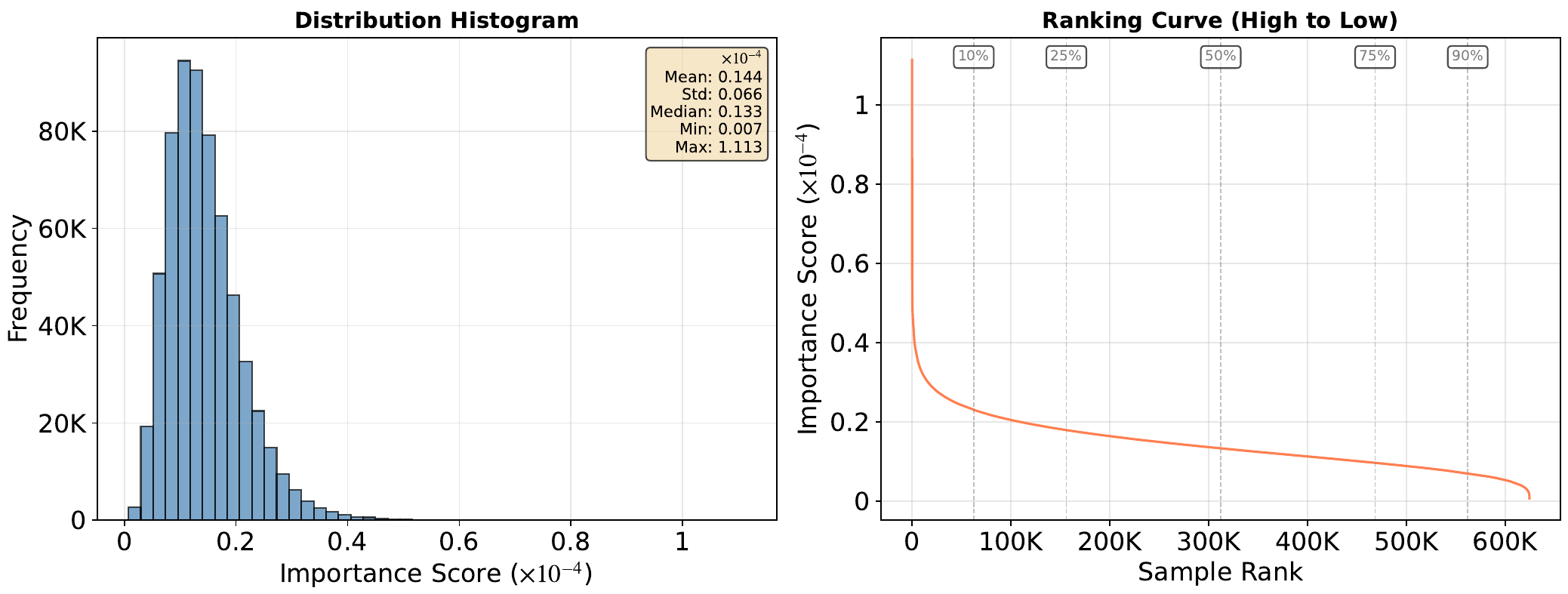}
   \caption{Distribution of Importance Scores. \textbf{Left:} Histogram of importance scores, exhibiting a highly skewed, long-tailed distribution.
    \textbf{Right:} Ranking curve of importance scores sorted in descending order, exhibiting a sharp initial drop followed by a long, gradually decaying tail.}
   \label{fig:distribution}
\end{figure}

\textbf{Probing the Distribution of Importance Scores.}
We further examine the distribution of the importance scores produced by ScalSelect.
As shown in Figure \ref{fig:distribution}, the scores exhibit a highly skewed, long-tailed distribution: a small fraction of samples contributes disproportionately to the dominant representation subspace, while the majority of samples have negligible leverage.
In addition, the ranking curve exhibits a sharp initial drop followed by a long, gradually decaying tail, indicating that a few top-ranked samples dominate the contribution, while importance decreases steadily over the remaining samples.
This suggests that ScalSelect supports stable and robust selection across a wide range of selection budgets.
This observation is consistent with the empirical trends reported in Section \ref{sec:main_res}.

\section{Conclusion}
We propose ScalSelect, a scalable and training-free multimodal data selection method for efficient visual instruction tuning.
ScalSelect leverages instruction-conditioned early representations and subspace-aware global selection to preserve the dominant structure of the full dataset without pairwise comparisons or auxiliary models.
Extensive experiments across multiple models, datasets, and selection budgets demonstrate that ScalSelect retains most of the performance of full-data training using only a small fraction of the data, and even surpasses it in some settings.
We believe ScalSelect offers a practical and principled solution for large-scale multimodal data selection and opens new directions for subspace-oriented data selection in vision-language learning.
\clearpage


\bibliographystyle{ACM-Reference-Format}
\bibliography{sample-base}

\appendix
\section{Experiment Details}\label{sec:details}
\subsection{Settings of Extracting Representation}
For Instruction-Conditioned Early Representation, we perform inference using the target VLM with a maximum context length (\texttt{max\_len}) of 4096 tokens. For LLaVA-Vicuna-7B, input images are processed following the default configuration of LLaVA-1.5, with a fixed resolution of 336 $\times$ 336. For Qwen3-VL-4B-Instruct and Qwen3-VL-8B-Instruct, input images are resized such that the maximum number of visual tokens does not exceed 576, with a patch resolution of 32 $\times$ 32.

\subsection{Training Hyperparameters}
All experiments are conducted on 8 $\times$ NVIDIA H100 GPUs (80GB) using LLaMA-Factory. The detailed training hyperparameters are summarized in Table \ref{tab:hyperparam}. The same training configuration is used across all models, datasets, and selection budgets to ensure fair comparison.

\begin{table}[!htbp]
\centering
\renewcommand{\arraystretch}{1.2}
\setlength\tabcolsep{8pt} 
\resizebox{0.6\linewidth}{!}
{
\begin{tabular}{l|c}
\toprule
\textbf{Hyperparameter} & \textbf{Value} \\
\midrule
Batch size per device & 4   \\
Gradient accumulation steps & 2 \\
Learning rate (LR) & 2e-5 \\
LR schedule & Cosine decay \\
LR warmup ratio & 0.03 \\
Weight decay & 0 \\
Epoch & 1 \\
DeepSpeed stage & 3\\
Cut length & 4096\\
\bottomrule
\end{tabular}
}
\caption{Training Hyperparameters.}
\label{tab:hyperparam}
\end{table}

\subsection{Settings of Baselines}

All baseline methods included in our experiments are applied by following the original experimental settings and implementation details described in their respective papers.

\section{Sensitivity of the Attention-Mass Threshold}\label{sec:threshold}

We further analyze the effect of the attention-mass threshold $\tau$ used in the Instruction-Conditioned Early Representation.
Table \ref{tab:threshold} reports the average retained visual token ratio and the overall relative performance under different values of $\tau \in \{0.85, 0.90, 0.95\}$, conducted on LLaVA-Vicuna-7B using the LLaVA-V-625K dataset.

As $\tau$ increases, the proportion of retained visual tokens grows gradually, resulting in diminishing marginal gains in token reduction.
Meanwhile, the overall relative performance remains stable across this range, with only minor variations.
These results indicate that the attention-mass threshold primarily affects representation compactness rather than introducing strong performance sensitivity.
Based on this observation, we fix $\tau = 0.9$ as a reasonable default choice that balances visual coverage and representation compactness throughout this work.

\begin{table}[!bhtp]
\centering
\renewcommand{\arraystretch}{1.2}
\setlength\tabcolsep{8pt} 
\resizebox{0.6\linewidth}{!}
{
\begin{tabular}{l|c|c}
\toprule
\textbf{$\tau$} & \textbf{Retained visual token ratio} & \textbf{Rel.} \\
\midrule
0.85 & 77.8\% & 99.26\% \\
\rowcolor{navyblue!10}0.90 & 84.6\% & 100.00\% \\
0.95 & 92.0\% &  99.33\% \\
\bottomrule
\end{tabular}
}
\caption{Retained visual token ratio and overall relative performance under different attention-mass thresholds $\tau$.}
\label{tab:threshold}
\end{table}

\section{Subspace Approximation Reference for ScalSelect}
We analyze the subspace approximation property of the selected subset in ScalSelect.
All notations follow Section \ref{sec:method} of the main paper.

\hypbox{Proposition (Relative-Error Subspace Approximation, Sampling Reference)}{Let $\mathbf{X}_k$ denote the best rank-$k$ approximation of the full representation matrix $\mathbf{X}_c$.
Let $\mathbf{X}_S$ be the submatrix formed by the samples selected according to leverage scores.
If samples are selected via randomized leverage-score sampling, then, with probability at least $99\%$:
\begin{equation}
\| \mathbf{X}_c - P_{\mathbf{X}_S} \mathbf{X}_c \|_F \;\le\; (1 + \varepsilon)\, \| \mathbf{X}_c - \mathbf{X}_k \|_F,
\label{eq:approxi}
\end{equation}
where $P_{\mathbf{X}_S}$ denotes the orthogonal projection onto the row space of $\mathbf{X}_S$, and $\varepsilon \in (0,1)$ is an accuracy parameter controlling the approximation slack.
}

\textbf{Proof.}
We adapt the relative-error analysis for leverage-score (subspace) sampling in CUR/CX decompositions \cite{drineas2008relative,mahoney2009cur}.
Let $A = X_c^\top \in \mathbb{R}^{d \times N}$.
Selecting a subset of samples (rows) from $X_c$ is equivalent to selecting the corresponding
subset of columns from $A$.

Let $A_k$ be the best rank-$k$ approximation of $A$, and let $P_{\mathcal{C}}$ denote the orthogonal projector onto the span of the sampled columns of $A$ (equivalently, the span of the sampled rows of $X_c$).
By leverage-score sampling (i.e., sampling proportional to the squared $\ell_2$-norms of the rows of the top-$k$ right singular vector matrix of $A$), the relative-error CX guarantee states that, for any accuracy parameter $\varepsilon \in (0,1)$, sampling a sufficiently large number of columns (e.g., with the standard $O(k\log k/\varepsilon^2)$ oversampling rate) yields, with probability at least $99\%$~\cite{drineas2008relative,mahoney2009cur}:
\begin{equation}
\|A - P_{\mathcal{C}}A\|_F \le (1+\varepsilon)\,\|A - A_k\|_F .
\end{equation}

Finally, mapping back to our notation, the sampled columns of $A=X_c^\top$ correspond to
the sampled rows forming $X_S$, and $P_{\mathcal{C}}$ corresponds to $P_{X_S}$.
Using the invariance of the Frobenius norm under transposition, we obtain
\begin{equation}
\begin{aligned}
\|X_c - P_{X_S}X_c\|_F
&= \|A^\top - (P_{\mathcal{C}}A)^\top\|_F \\
&= \|A - P_{\mathcal{C}}A\|_F \\
&\le (1+\varepsilon)\,\|A - A_k\|_F\\
&= (1+\varepsilon)\,\|X_c - X_k\|_F ,
\end{aligned}
\end{equation}
which completes the proof. \qed

\paragraph{Remark on Deterministic Selection.}
The relative-error bound stated above follows the classical analysis of \emph{randomized} leverage-score sampling in CUR/CX decompositions \cite{drineas2008relative,mahoney2009cur}, and serves as a theoretical reference for the design of ScalSelect.
In practice, leverage scores are commonly used as a \emph{deterministic importance measure}, and we accordingly adopt a deterministic top-leverage selection strategy by selecting the highest-scoring samples, a well-established heuristic motivated by leverage-score sampling theory that improves stability by removing randomness \cite{mahoney2011randomized,drineas2008relative}.
The above result provides theoretical motivation for using leverage scores as a principled criterion for sample selection.


\end{document}